\documentclass[journal]{IEEEtran}
\IEEEoverridecommandlockouts
\usepackage{cite}
\usepackage{bm}
\usepackage{amsmath,amssymb,amsfonts}
\DeclareMathOperator*{\argmax}{arg\,max}
\usepackage{graphicx}
\usepackage{textcomp}
\usepackage{booktabs}
\usepackage{xcolor}
\def\BibTeX{{\rm B\kern-.05em{\sc i\kern-.025em b}\kern-.08em
    T\kern-.1667em\lower.7ex\hbox{E}\kern-.125emX}}

\usepackage{mwe}
\usepackage{mathtools}
\usepackage{amssymb}
\usepackage{subfiles}
\usepackage{collectbox}
\usepackage{multicol}
\usepackage{xcolor} % http://ctan.org/pkg/xcolor
\usepackage[boxed, ruled, linesnumbered]{algorithm2e}

\SetCommentSty{mycommfont}
\usepackage{svg}

\usepackage{float}
\usepackage{graphicx}

\usepackage{times}
\usepackage{soul}
\usepackage{url}
\usepackage{hyperref}
\usepackage[utf8]{inputenc}
\usepackage{caption}
\usepackage[colorinlistoftodos]{todonotes}
\usepackage{comment}

\title{A Comparison of Self-Play Algorithms Under a Generalized Framework
\thanks{This work was funded by the EPSRC Centre for Doctoral Training in Intelligent Games and Game Intelligence (IGGI) EP/L015846/1.}}
\author{
\IEEEauthorblockN{Daniel Hernandez\IEEEauthorrefmark{1}, Kevin Denamganaï\IEEEauthorrefmark{1}, Sam Devlin\IEEEauthorrefmark{2},\\ Spyridon Samothrakis\IEEEauthorrefmark{3} and James Alfred Walker\IEEEauthorrefmark{1},~\IEEEmembership{Senior Member, IEEE}}\\

\IEEEauthorblockA{\IEEEauthorrefmark{1}Department of Computer Science, University of York, UK.~\{dh1135, kyd500, pjy500, james.walker\}@york.ac.uk}

\IEEEauthorblockA{\IEEEauthorrefmark{2}Microsoft Research, Cambridge, UK.~sam.devlin@microsoft.com}

\IEEEauthorblockA{\IEEEauthorrefmark{3}Institute of Analytics \& Data Science, University of Essex, UK.~ssamot@essex.ac.uk}
}

\begin{document}

\IEEEpubid{\begin{minipage}{\textwidth}\ \\[12pt]
        978-1-7281-1884-0/19/\$31.00 \copyright 2019 IEEE
\end{minipage}}

\maketitle

\begin{abstract}
    Throughout scientific history, overarching theoretical frameworks have allowed researchers to grow beyond personal intuitions and culturally biased theories. They allow to verify and replicate existing findings, and to link disconnected results. The notion of self-play, albeit often cited in multiagent Reinforcement Learning, has never been grounded in a formal model. We present a formalized framework, with clearly defined assumptions, which encapsulates the meaning of self-play as abstracted from various existing self-play algorithms. This framework is framed as an approximation to a theoretical solution concept for multiagent training. On a simple environment, we qualitatively measure how well a subset of the captured self-play methods approximate this solution when paired with the famous PPO algorithm. We also provide insights on interpreting quantitative metrics of performance for self-play training. Our results indicate that, throughout training, various self-play definitions exhibit cyclic policy evolutions.
\end{abstract}

\vspace{-1em}

\section{Introduction}\label{section:introduction}

% TODO
% Name Nash averaging and alpharank (with alpha alpharank) as plausible evaluation metrics. Speak a bit about tractability.

In the classical single agent reinforcement learning (RL) scenarios described by~\cite{Sutton1998}, where a stationary environment is modelled by a Markov Decision Process (MDP), a solution concept can be defined. MDPs are solved by computing a policy which yields the highest possible episodic reward. However, it is not clear how to define a pragmatic solution concept when training a single policy in a multi-agent system, for an agent's optimal strategy is dependent on behaviours of the other agents that inhabit the environment. An initial solution is to compute the expected reward obtained by a given policy defined over the \textit{entire} set of all possible other policies in the environment, which is intractable in all but toy scenarios.

To approximate this solution, traditional multi-agent RL (MARL) methods would train and benchmark a policy against a set of preexisting fixed agents, using as a success metric the relative performance against these agents. These methods rest on two assumptions. Firstly, the availability of benchmarking policies at training and testing time. Secondly, these existing policies dominate, in a game theoretical sense, most of the policy space. Thus it would not be necessary to compute the expectation over the entire policy space, using as a proxy an expectation over the preexisting policies.

However, this approach features many flaws. if this benchmarking set of is too small, the trained policy may overfit to the behaviour of the agents it was trained with, and thus prone to being exploitable by other policies. Furthermore, the validity of the last assumption is rarely formally justified, favouring empirical results. 

% Within the field of RL there are multiple methods for computing these benchmarking policies which must be available before training commences. To name a few, these preexisting policies can be computed using supervised learning on datasets of expert human moves to bias learning a policy towards expert human play~\cite{silver2016mastering}\cite{Tesauro1992}; they can be tree-search based algorithms using hand-crafted evaluation functions or Monte Carlo based approaches if an environment model is present~\cite{browne2012survey}. Some methods are as creative as deriving a strong policy by using off-policy methods on video replays~\cite{Aytar2018}\cite{Malysheva2018}.

What about the cases in which we don't have access to these learning resources? Such as when developing a new game for which no prior expert information is known, and for which any hand-crafted evaluation functions yields a fruitless policy. A priori methods such as optimistic policy initialization are still permitted~\cite{Machado2014}. Yet, under such constraints, there is little room to compute a set of good benchmarking policies, let alone a set of dominating policies.%policies which we can assume dominates the majority of the policy space.

Authors such as~\cite{Samuel1959} began experimenting on self-play (SP). SP is an open-ended learning training scheme which arises in the context of multi-agent training. A SP training scheme trains a learning agent \textit{purely} by simulating plays with itself, or with policies which have been generated during training. These generated policies can dynamically build a set of benchmarking policies during training. Such set can potentially be curated to remove dominated or redundant policies.

Once we leave behind the limiting approach of training against a fixed and known set of policies in favour of SP, it is of paramount importance to define meaningful metrics to inform this open-ended learning process. Fortunately, recent years have seen the introduction of metrics for multiagent evaluation, stemming from game theory~\cite{Balduzzi2019} or dynamical systems analysis~\cite{omidshafiei2019alpha}.

Historically, SP lacks a formal definition, and notation is often not shared among researchers. This has led to isolated, and sometimes conflicting, conceptions of what constitutes SP as a training scheme in MARL\@. It is our firm belief that a formally-grounded framework with rigorous and unified notation will strengthen the field of SP MARL and allow for the creation of more nuanced and efficient contributions. Incremental efforts on existing and future contributions can now be captured on a shared language. This paper constitutes a first step towards defining a generalizing framework under which SP MARL methods can be inspected. Our contributions:

% Self-play decides which policies will define the behaviour of other agents in the environment, by choosing a set of fixed strategies to play against, we are essentially fixing a Markov Game for the agent under policy $\pi$ to act on. The SP module presents the learning agent $\pi$ with a sequence of Markov Games throughout the duration of the training.

\begin{itemize}
    \item A generalizing framework defined under formal notation to describe SP algorithms in MARL\@.
    \item A unifying definition under the presented framework of some prevalent SP algorithms from the literature.
    \item A qualitative and quantitative study of some SP algorithms.
\end{itemize}

\section{Related Work}\label{section:related-work}

The notion of SP has been present in the game playing AI community for over
half a century.~\cite{Samuel1959} discusses the notion of learning a
state-value function to evaluate board positions in the game of checkers, to
later inform a 1-ply tree search algorithm to traverse more effectively the
search space. This learning process takes place as the opponent uses the same
state-value function, both playing agents updating simultaneously the shared
state-value function. Such training fashion was named self-play. The TD-Gammon
algorithm~\cite{TDGammon} featured SP to learn a policy using
TD($\lambda$)~\cite{Sutton1998} to reach expert level backgammon play. This
approach surpassed previous work by the same author, which derived a backgammon
playing policy by performing supervised learning on expert
datasets~\cite{Tesauro1990}. More recently, AlphaGo~\cite{Silver2017a} used a
combination of supervised learning on expert moves and SP to beat the world
champion Go player. This algorithm was later refined~\cite{Silver2017b},
removing the need for expert human moves. A policy was learnt purely by using
an mix of supervised learning on moves generated by SP and Monte Carlo Tree
Search (MCTS), as presented in~\cite{Anthony2017}. These works echo the
sentiment that superhuman AI needs not be limited or biased by preexisting
human knowledge.

% In the game of Othello,~\cite{VanDerRee2013} experimented with training single
% agent RL algorithms using two different training schemes: SP and training
% versus a fixed opponent. Their results show that, depending on the RL algorithm
% used, learning by SP yields a higher quality policy than learning against a
% fixed opponent. Concretely, TD($\lambda$) learnt best from self-play, but
% Q-learning performed better when learning against a fixed opponent.
% Similarly,~\cite{Firoiu2017} found that deep Q-Network (DQN)~\cite{Mnih2013}, a
% deep variant of Q-learning, did not perform well when trained against other
% policies which were themselves being updated simultaneously, but otherwise
% performed well when training against fixed opponents in the fighting game
% \textit{Super Smash Brothers: Melee}. We note that that the environments in
% their experiments differ too much to draw parallel conclusions.

It is often assumed that a training scheme can be defined as SP if, and only
if, all agents in an environment follow the same policy, corresponding to the
latest version of the policy being trained. Meaning that, when the learning
agent's policy is updated, every single agent in the environment mirrors this
policy update. We refer to this SP method as \textit{naive}
SP.~\cite{Bansal2017} relaxes this assumption by allowing some agents to follow
the policies of ``past-selves''. Instead of replicating the same policy over
all agents, the policy of all of the non-training agents can \textit{also} come
from a set of \textit{fixed} ``historical'' policies. This set is built as
training progresses, by taking \textit{checkpoints}\footnote{For deep RL, this
is equivalent to freezing the weights of the neural networks to represent an
agent's policy.} of the policy being trained. At the beginning of a training
episode, policies are uniformly sampled from this ``historical'' policy set and
define the behaviour of some of the environment's agents. The authors claim
that such version of SP aims at training a policy which is able to defeat
random older versions of itself, ensuring continual learning. This notion of a
``choosing policies from a historical set'' allows for two decision points: (1)
Which agents will be added into this ``historical'' set of policies and (2)
which of these agents will populate the environment. Different takes on (1) and
(2) spawn different SP algorithms.

From this scenario, consider the following: each \textit{combination} of fixed
policies sampled as opponents from the ``historical'' dataset can be considered
as a separate MDP\@. This is because by leaving a single agent learning in a
stationary environment, the fixed agents' influence on the environment is
stationary~\cite{Laurent2011}. This is of genuine importance, given that most
RL algorithms' convergence properties heavily rely on the assumption of a
stationary environment~\cite{Asai2001}. SP algorithms can leverage the
assumption that they are using SP, so they can provide the learning agent with
a label denoting which combination of agent behaviours inhabits the
environment, a powerful assumption in transfer learning~\cite{Sutton2007} and
multi-task learning~\cite{Taylor2009}. In fact, there already are multitask
meta-RL algorithms which assume knowledge of a distribution over MDPs which the
agent is being trained on, such as RL$^2$~\cite{Duan2016}. Note that a SP
algorithm featuring a growing set of ``historical'' policies will introduce a
non-stationary distribution over the policies that will inhabit the environment
during training. It ensues that the distribution over the set of MDPs\@ encountered
by the training agent becomes non-stationary.

Recently,~\cite{openai2019dota} trained a team of RL agents using SP to achieve
superhuman level performance in the competitive team-based game of Dota 2.
During training, the team would play $80\%$ of the games using \textit{naive}
SP while the remaining $20\%$ were played against ``past-selves''. The
probability of facing any of these previous policies depends on a per-policy
metric (which is updated during training) evaluating how much is there to learn
from a policy. AlphaStar~\cite{vinyals2019grandmaster} reached Grandmaster
level in StarCraft II with various policies by using a combination of various
SP algorithms~\cite{Jaderberg2018}. Part of their training pipeline relied on
training a set of ``exploiter'' policies which focus on exploitining specific
policies under training, relaxing the need for them to be robust to all
opponents.

%a training pipeline relying on
%\textit{naive} SP ($35\%$) and population based training~\cite{Jaderberg2018},
%fictitious SP~\cite{heinrich2015fictitious}, league training ($65\%$) a
%\textit{prioritized} Fictitious Self-Play training scheme that makes use of
%``past-selves'' policies and so-called ``league exploiters'', i.e. policies
%which are trained alongside the main training policy to exploit specific
%policies in the league, relaxing the need for them to be robust to all
%opponents.

% (If there's space: refs 20,21 and 22 in alphastar paper) 

\cite{Lanctot2017} Defines the Policy-Space Response Oracles (PSRO) family of
algorithms, unifying various game theoretical algorithms for multiagent
training. PSRO algorithms tackle this problem by iteratively generating
monotonically stronger policies relative to an existing set of policies. These
algorithms iterate over the following loop: a meta-game (definition in
Section~\ref{section:preliminary_notation}) is defined over the current set of
policies, for which a ``solution`` is computed, and from this solution one or
more policies are added to the set of policies. The choice of solution concept
and the procedure to generate new policies from this concept is the
differentiating factor between PSRO algorithms. There are current efforts to
show convergence properties of some PSRO
algorithms~\cite{muller2019generalized}\cite{Balduzzi2019} towards existing
multiagent solutions~\cite{omidshafiei2019alpha}. Our contribution shares the
spirit of creating a generalised framework to encompass existing algorithms,
but with a focus on MARL literature instead of game theory.

% Unfortunately, the numerous empirical successes which motivate SP as a
% promising training scheme suffer from lack of formal proofs of convergence or
% rate thereof~\cite{Tesauro1992}. We hope to provide a simple, yet powerful tool
% to analyze SP schemes in the next section.

% TODO:
% Find where to add ``a generalized training approach for multiagent learning''
% Find where to add AlphaStar

\section{Preliminary notation}\label{section:preliminary_notation}

\vspace{-0.5em}
\noindent
Cursive lowercase letters represent scalars ($n$). Bold lowercase, vectors ($\bm{\pi} \in \mathbb{R}^n$). Bold uppercase, matrices ($\bm{A} \in \mathbb{R}^{n \times n}$).
\vspace{-1.3em}

\subsection{Normal form games}

A \textbf{normal form game} is a tuple ($\Pi$, $U$, $n$) where $n$ is the number of players, $\Pi = (\Pi_1, \ldots, \Pi_n)$ is the set of joint policies, one for each player. $U: \Pi \rightarrow \mathbb{R}^n$ is a payoff table mapping each joint policy to a scalar utility for each player.

Rational players try to maximize their own expected utility. Each player  $i$ does so by selecting a policy from $\Pi_i$ or equivalently by sampling from a mixture (distribution) over them $\pi_i \in \Delta{}(\Pi_i)$. The \textbf{value} $v_i$ for a player $i$ given a policy vector $\bm{\pi}$ is the expected payoff obtained by player $i$ if all players follow $\bm{\pi}$, $v_i = U_{i}(\bm{\pi})$. 

A (possibly mixed) policy $\pi_i$ is a best response for player $i$ against all other players' policies $\bm{\pi_{-i}}$ if playing $\pi_i$ yields player $i$ the highest possible payoff against strategies $\bm{\pi_{-i}}$, $\pi_i \in BR(\bm{\pi_{-i}})$. A Nash Equilibrium is a policy profile (one policy for each player) such that each player's policy is a best response against all other player policies. $\forall i \in \{n\}, \; \pi_i \in BR(\bm{\pi_{-i}})$. A Nash Equilibrium is maximally entropic (maxent Nash) if each player's policy is maximally indifferent between actions with the same empirical performance. 

% Do wo say this?
%\cite{Balduzzi2018} shows that symmetrical 2-player zero-sum games with an antisymmetric\footnote{A matrix $\bm{A}$ is antisymmetric iff: $ -\bm{A} = \bm{A}^T$}payoff matrix $\bm{A}$ have a unique maxent Nash. To compute a maxent Nash we use the algorithm presented in~\cite{ortiz2007maximum}.

A game is zero-sum if $\forall \pi \in \Pi, \; \bm{1} \cdot U(\Pi) = 0$, otherwise it is a general-sum game. A game is symmetric if all players feature the same policy set ($\Pi_1 = \ldots = \Pi_n$) and the payoff associated to each joint policy depends only on the policies and not on the identity of the players. 2 player normal form games ($n = 2$) are typically defined by a tuple ($\bm{A}$, $\bm{B}$), where $\bm{A} \in \mathbb{R}^{|\Pi_1| \times |\Pi_2|}$ gives the payoff for player 1 (row player), and $\bm{B} \in \mathbb{R}^{|\Pi_1| \times |\Pi_2|}$ gives the payoff for player 2 (column player). If $\bm{B} = \bm{A^T}$ the game is symmetric. Most importantly for us, if $\bm{B} = -\bm{A}$ the game is zero-sum. Exploiting this equality, 2-player zero-sum games are often represented by a single matrix $\bm{A}$ containing the payoffs for player 1.

Given a vector of $n$ agents $\bm{\pi}$ for an arbitrary game, also known as a population, let $\bm{W}_{\bm{\pi}} \in \mathbb{R}^{n \times n}$ denote an \textbf{empirical winrate matrix} also known as a \textbf{meta-game}. The entry $w_{i,j}$ for $i,j \in \{n\}$ represents the winrate of many head-to-head matches of policy $\pi_i$ when playing against policy $\pi_j$ for the given game. A meta-game can be thought of as an abstraction of the underlying game, in which a players' actions consist of choosing policies from the population rather than primitive game actions. A meta-game's empirical winrate matrix $\bm{W_{\pi}}$ for a given population $\bm{\pi}$ can be considered as a payoff matrix for a 2-player zero-sum game. It is possible to define an empirical winrate matrix over two (or more) populations $\bm{W}_{\bm{\pi_1}, \bm{\pi_2}}$, such that each player chooses agents from a different population. An \textbf{evaluation matrix}~\cite{Balduzzi2019} is a meta-game represented by an antisymmetric matrix $\bm{A}$. One can turn an empirical winrate matrix $\bm{W}$ into an antisymmetric matrix by performing the element wise operation $a_{i,j} = w_{i, j} - \frac{1}{2}$, shifting the range of each entry from $[0, 1]$ to $[-\frac{1}{2}, \frac{1}{2}]$. Symmetrical 2-player zero-sum games represented by an antisymmetric matrix $\bm{A}$ feature a unique maxent Nash~\cite{Balduzzi2019}, a fact we will use in Section~\ref{section:examples-from-literature}.

Finally, the \textbf{relative population performance}~\cite{Balduzzi2019} is a population-level meassure of performance. Given two populations $\bm{\pi_1}, \bm{\pi_2}$, it yields a single scalar value comparing the performance of $\bm{\pi_1}$ against $\bm{\pi_2}$. It is computed by generating an evaluation matrix for both populations $A_{\bm{\pi_1}, \bm{\pi_2}}$ which is then treated as a 2-player zero-sum game. A Nash equilibrium is then computed $(\bm{n_{\pi_1}}, \bm{n_{\pi_2}})$ for the zero-sum game defined by $A_{\bm{\pi_1}, \bm{\pi_2}}$. The relative population performance is the value $v$ for the meta-player 1: $v = \bm{n_{\pi_1}} \cdot A_{\bm{\pi_1}, \bm{\pi_2}} \cdot \bm{n_{\pi_2}}^T$. A positive $v$ indicates that $\bm{\pi_1}$ wins on average against population $\bm{\pi_2}$, with the opposite being true if $v$ is negative, $v=0$ indicates both populations are equivalent.

\subsection{Multiagent Reinforcement Learning}

Let \textit{E} represent a multi-agent system with $n$ agents and a reward discount factor $\gamma$. This environment \textit{E} features a state space $S$, a joint observation space $O = O_1 \times \ldots \times O_n$ and a joint action space $A = A_1 \times \ldots \times A_n$, where $O_i$ and $A_i$ represent the observation and action space for the $i$th agent respectively. Let the (potentially stochastic) mapping from observations to actions $\pi_i: O_i \rightarrow A_i$ represent the policy for the $i$th agent, and $\boldsymbol{\pi} = [\pi_1,\ldots,\pi_n]$ the joint policy vector, containing the policy for each agent in \textit{E}. The joint policy vector $\boldsymbol{\pi}$ can also be regarded as a distribution over the joint action space conditioned on the joint observation space $\boldsymbol{\pi}: O \rightarrow A$. Let $\Pi = \Pi_1 \times \ldots \times \Pi_n$ be the joint policy space, where $\Pi_i$ is the policy space for agent $i$. As before, let $\Pi_{-i}$ denote the joint policy space for all agents except agent $i$.

The solution to this environment \textit{E} for an agent $i$ is to compute a policy which maximizes its expected reward obtained when acting in an environment across the \textit{entire} set of all possible other policies $\Pi_{-i}$ in the environment:

\begin{equation}
    \pi^* = \argmax_{\pi \in \Pi_i} \int_{\boldsymbol{\pi_{-i}} \subseteq \Pi_{-i}} \mathbb{E}_{\boldsymbol{a_t} \sim \boldsymbol{\pi}; s_{t+1}, r_t \sim P(s_t,  \boldsymbol{a_t})}[\sum_{t=0}^\infty \gamma^t r_t]
    \label{equation:multi-agent-solution}
\end{equation}

An iteration, or episode, of the classical MARL loop goes as follows: The environment presents all agents with a vector containing all individual agent observations $\boldsymbol{o_t} = [o^1_t, \ldots, o^n_t]$ based on its state $s_t$. The vector containing the actions of all agents is sampled from the joint policy vector $\boldsymbol{a_t} \sim \boldsymbol{\pi}(\boldsymbol{o_t})$. The environment then executes the action vector $\boldsymbol{a_t}$, transitioning to a new state $s_{t+1}$ and yielding both a new observation $\boldsymbol{o_{t+1}}$ and a reward vector $\boldsymbol{r_t}$ containing an observation and reward for each agent. This loop is repeated until a terminal state is reached, after which a new episode begins.

\section{Generalized Self-Play Framework}\label{section:self-play-framework}

Here we present the mathematical formulation, and required assumptions, for a formal framework which encapsulates the notion of self-play in the context of MARL\@. It allows for the creation and comparison of existing and future SP algorithms.

Self-play training schemes can be conceived as modules which extend the MARL loop by introducing a functionality prior to, and after, every episode. Let $\pi$ be the only policy being trained throughout the MARL loop. An SP scheme envelops the MARL loop by first deciding which policies $\boldsymbol{\pi'}$, taken from a set of \textit{fixed} policies $\boldsymbol{\pi'} \subseteq \boldsymbol{\pi^o}$, will define the agents' behaviour for the next episode. This \textit{excludes} the agent whose behaviour is defined by $\pi$. Once the episode ends, a function $G$ decides whether or not the (possibly updated) policy $\pi$ will be introduced in the pool of available policies $\boldsymbol{\pi^o}$. This intuition is formally captured in Algorithm~\ref{algorithm:posg-self-play-rl-loop}, which presents a SP scheme inside a Partially Observable Stochastic Game (POSG) loop. Algorithm~\ref{algorithm:posg-self-play-rl-loop} defines an \textit{n-player}, \textit{general-sum}, \textit{partially-observable} environment. The steps belonging to the SP scheme have been highlighted in orange.

\begin{algorithm}
    \KwIn{ \textit{Environment}: $(S, A, O, \mathcal{P}(\cdot, \cdot| \cdot, \cdot), \mathcal{R}(\cdot, \cdot), \rho_0)$}
    \KwIn{ \textcolor{orange}{ \textit{Self-Play Scheme}: $(\Omega(\cdot | \cdot, \cdot), G(\cdot | \cdot, \cdot))$}}
    \KwIn{ \textit{Policy to be trained}: $\pi \in \Pi_i$}
    \textcolor{orange}{$\boldsymbol{\pi^o} = \{ \pi \}$} \tcp*{Menagerie initialization}
    \For{$e = 0,1,2,\ldots\;$} {
        \textcolor{orange}{$\boldsymbol{\pi'} \sim \Omega(\boldsymbol{\pi^o}, \pi)$} \tcp*{Sample from menagerie}
        \textcolor{orange}{$\boldsymbol{\pi} = \boldsymbol{\pi'} \cup \{\pi\}$\;}
        $s_0, \boldsymbol{o_0} \sim \rho_0$\;
        \For{$t = 0,\ldots, $ termination} {
            $\boldsymbol{a_t} \sim \boldsymbol{\pi}(\boldsymbol{o_t})$\;
            $s_{t+1}, \boldsymbol{o_{t+1}} \sim P(s_t, \boldsymbol{a_t})$\;
            $\boldsymbol{r_t} \sim \boldsymbol{R}(s_t, \boldsymbol{a_t})$\;
            $t \leftarrow t + 1$\;
        }
        $\pi \leftarrow update(\pi)$\;
        \textcolor{orange}{$\boldsymbol{\pi^o} \sim G(\boldsymbol{\pi^o}, \pi)$} \tcp*{Curate menagerie}
    }
    return $\pi$\;

    \caption{(POSG) RL Loop with Self-Play.}\label{algorithm:posg-self-play-rl-loop}
\end{algorithm}

\subsection{Framework definition}

We define a SP module or training scheme by formalizing the notions of the \textit{menagerie} $\boldsymbol{\pi^o}$, the \textit{policy sampling distribution} $\Omega$, and the \textit{gating function} $G$. Specified by the tuple $<\Omega(\cdot | \cdot, \cdot), G(\cdot | \cdot, \cdot)>$:
\begin{itemize}
    \item $\boldsymbol{\pi^o} \subseteq \Pi_i$; The \textbf{menagerie}. A set of policies from which agents' behaviour will be sampled. This set always includes the currently training policy $\pi$. A constraint is placed over $\boldsymbol{\pi^o}$. All of its elements must be derived, at least indirectly, from $\pi$, the policy being trained. Hence, all policies in the menagerie are elements of $\pi$'s policy space. The menagerie \textit{can} change as training progresses by the curator function described below. 

    \item $\Omega(\boldsymbol{\pi'} \in \Pi_{-i} | \boldsymbol{\pi^o} \subseteq \Pi_i, \pi \in \Pi_i) \in [0, 1]$; where $\boldsymbol{\pi'} \subseteq \boldsymbol{\pi^o}$; The \textbf{policy sampling distribution}. A probability distribution over the menagerie $\boldsymbol{\pi^o}$, the set of available policies. It is conditioned on the menagerie $\boldsymbol{\pi^o}$ and the current policy $\pi$ being trained. It chooses which policies, apart from $\pi$, will inhabit the environment's agents.

    \item $G(\boldsymbol{\pi^o}' \subseteq \Pi_i | \boldsymbol{\pi^o} \subseteq \Pi_i, \pi \in \Pi_i) \in [0,1]$; The \textbf{curator} or gating function, of the menagerie. A possibly stochastic function whose parameters are the current training policy $\pi$ and a menagerie $\boldsymbol{\pi^o}$. The curator serves two purposes, which complex curators could break into two functions:
        \begin{itemize}
            \item $G$ decides if the current policy $\pi$ will be introduced in the menagerie.
            \item $G$ decides which policies in the menagerie, $\pi \in \boldsymbol{\pi^o}$, will be discarded from the menagerie.
        \end{itemize}
\end{itemize}
The curator bears resemblance with the notion of Hall of Fame from evolutionary algorithms~\cite{Nogueira2013}. As Hall of Fame algorithms also consider the problem of curating a policy set over time.

% \todo[inline, color=red]{Kevin@Dani: in order to simplify the menagerie, shouldn't we define it on the side of the learning policy $\pi$ so that we do not have to constrain the sampling distribution with the requirement of always sampling it from the menagerie. Moreover, in the algorithms and the code, it does not work that way, the learning policy is clearly kept outside of the menagerie and added afterwards through the union operator (line 5). It does not prevent the initialization of the menagerie with a copy of the learning policy though. }

\subsection{Assumptions}\label{subsection:self-play-assumptions}

\noindent
Our SP framework explicitly assumes the following:

\noindent
\textbf{Assumption 1.1:}
\textit{The policies present in the environment can either be exact copies of the policy being trained, or policies derived indirectly from it, taken from the menagerie.}\\
\textbf{Assumption 1.2:}
\textit{Prior, during and after a training episode, the SP module has access to the agents' policy representations\footnote{If the policies are being represented by a neural network. Access to the policy representation means access to the neural network topology and weights.}. Allowing any-time read and write rights for all policies.}

The definitions above capture the minimal structure of all SP training schemes. However, it is possible to condition both the policy sampling distribution $\Omega$ and curator $G$ on any other variables. For instance, it could be interesting to define an SP algorithm whose components are conditioned on episode trajectories, which has proved useful in RL research~\cite{Schaul2015}, and is required for policy gradient algorithms~\cite{Williams1992}.

Our SP framework does not make any assumptions on the environment with which the policies interact.

% With MMDPs, dec-POMDPs and Partially Observable Stochastic Games being some of the most used environment models in the literature.

\subsection{Self-play as an approximation to the multiagent solution}

% SP as an approximate solution to MARL problems assumes: 
\textbf{Assumption 1:}
\textit{There exists a set of policies, $\boldsymbol{\pi} \subseteq \Pi$, significantly smaller than the entire original policy space, $|\boldsymbol{\pi}| \ll |\Pi|$, which we can use as a proxy for $\Pi$ in equation~\ref{equation:multi-agent-solution}. If so, the integration over the policy space, becomes computationally tractable. Making equation~\ref{equation:multi-agent-solution} computationally solvable.}

The policy sampling distribution $\Omega$ and the gating function $G$ are tools by which a menagerie $\boldsymbol{\pi^o}$ can be computed and curated over time. Self-play can be conceived as a bottom up approach towards computing a set of policies, $\boldsymbol{\pi^o}$, to be used as a proxy for the entire policy space $\Pi$ in equation~\ref{equation:multi-agent-solution}.
The obvious fact that an agent cannot act according to a policy outside its policy space means that a menagerie can only contain policies of a single policy space. Consequently, for environments with disjoint policy spaces, SP may be unable to serve as an approximate solution to equation~\ref{equation:multi-agent-solution}.

~\cite{Balduzzi2019} introduces the notion of the \textit{gamescape}, a polytope which geometrically encodes interactions between agents for zero-sum games. They derive a set of algorithms whose goal is to grow and curate an approximation to this polytope. We draw parallels between their work and the idea of using SP algorithms to compute a proxy for a target policy space.

\section{Self-Play Algorithms}\label{section:examples-from-literature}

We demonstrate the generalizing capabilities of our framework by presenting four prevalent SP schemes from MARL literature. Let $\pi$ be a policy being trained, and $\boldsymbol{\pi^o}$ a menagerie:

\subsubsection{Naive Self-Play}

The is the oldest and simplest SP algorithm, originating in~\cite{Samuel1959}. The premise is that every agent in the environment is populated with the latest version of the policy being trained. All agents share the same behaviour. To capture this, the policy sampling distribution $\Omega$ puts all probability weight to the latest $\pi$.

\begin{equation*}
    \Omega(\boldsymbol{\pi'} | \boldsymbol{\pi^o}, \pi) = 
    \begin{cases*}
        1 & $\forall \pi' \in \boldsymbol{\pi'}:\; \pi'$ == $\pi$ \\
        0 & otherwise
    \end{cases*}
\end{equation*}

In this degenerate scenario the gating function $G$ always deterministically inserts the latest version of the training policy into the menagerie, discarding the previous menagerie entirely. 
\begin{equation*}
    G(\boldsymbol{\pi^o}, \pi) = \{\pi \}
\end{equation*}

\subsubsection{$\delta$-Uniform Self-Play}

Introduced by~\cite{Bansal2017} and mentioned in
Section~\ref{section:related-work}. This SP scheme treats the menagerie as a set of
``historical'' policies. The authors wanted to create an SP scheme that
ensured continual learning by training a policy which could consistently beat
random older versions of itself.

Let $M = |\boldsymbol{\pi^o}|$ be the size of the menagerie, and let $\delta
\in [0, 1]$ denote the percentage threshold on the oldest policy to be 
considered as a potential candidate to be sampled from $\bm{\pi^o}$ by
$\Omega$. Thus, $\delta = 0$ corresponds to all policies in the menagerie being
considered as candidates, and $\delta = 1$ only allows the last policy
introduced in the menagerie to be sampled by $\Omega$. After computing the set
of candidate policies following this criteria, the authors use a uniform
distribution to sample from it.

\begin{equation*}
    \Omega(\boldsymbol{\pi'} | \boldsymbol{\pi^o}, \pi) = Uniform(\delta M, M)
    \label{equation:delta-uniform-self-play-omega}
\end{equation*}

The gating function $G$ used in $\delta$-uniform-self-play is fully inclusive
and deterministic. After every episode, it always inserts the training policy
into the menagerie.

\begin{equation*}
    G(\boldsymbol{\pi^o}, \pi) = \boldsymbol{\pi^o} \cup \{\pi\}
\end{equation*}

\subsubsection{Population Based Training Self-Play}

As introduced in~\cite{Jaderberg2018}, Population Based Training SP is a
parallel SP algorithm influenced by evolutionary algorithms.
Each agent is independently learning on their own SP augmented MARL loop. The
menagerie, initialized with a population of random policies, is shared amongst
all learning agents. The menagerie is treated as the population of an
evolutionary algorithm.

The policy sampling distribution chooses opponents from the menagerie which are
similar in skill to the currently training agent. Where agent skill is
meassured by Elo ratings.

The gating function is analogous to the selection, crossover and mutation
phases of an evolutionary algorithm. It modifies and changes the menagerie by
dropping low performing agents and introducing evolved versions of the existing
population.

\subsubsection{Policy-Spaced Response Oracles (PSRO)}

%- Update payoff tensor M for new policy profiles in menagerie via game simulatinos.
%- Compute meta-strategy using meta-solver $\mathcal{M}(M)$
%- Expand policy space via Oracle

A family of algorithms introduced in \cite{Lanctot2017}. Such algorithms maintain an empirical winrate matrix $\bm{W_{\pi^o}}$ generated from a menagerie $\bm{\pi^o}$, and are parameterized via the choice of two functions:
\begin{itemize}
    \item $\mathcal{M}(\bm{W_{\pi^o}} \in \mathbb{R}^{|\bm{\pi^o}| \times |\bm{\pi^o}|}) \in \Delta(\bm{\pi^o})$. The meta-game solver, which takes a meta-game and outputs a ``meta-game solution'', a distribution over the policies of the menagerie. 
    \item $\mathcal{O}(\pi \in \Pi, \bm{\pi'} \in \Delta(\bm{\pi^o})) \in \Pi$. The oracle, which takes a distribution over policies $\bm{\pi'}$, a starting policy $\pi$ and derives a new policy $\pi^*$ which performs better against $\bm{\pi'}$ than $\pi$.
\end{itemize}

The function of the meta-game solver $\mathcal{M}$ is captured by our policy sampling distribution $\Omega$, as they both output a probability distribution over a set of policies, the menagerie. After the oracle computes a new policy, it is added to the meta-game, and the empirical winrate matrix $\bm{W_{\pi^o}}$ is updated via game simulations.

$\mathcal{M}$ operates on a meta-game generated by doing head-to-head matches between all policies in the menagerie, whereas a policy sampling distribution $\Omega$ operates directly on the menagerie. In this paper we use $\mathcal{M} =$ maxent-Nash~\cite{Balduzzi2018}. As stated in Section~\ref{section:preliminary_notation}, we can turn a winrate matrix into an antisymmetric evaluation matrix, which we know has a unique maxent Nash. This uniqueness feature is valuable for consistent interpretability. Other alternatives exist \cite{omidshafiei2019alpha}\cite{muller2019generalized}.

\begin{equation*}
    \Omega(\boldsymbol{\pi'} | \boldsymbol{\pi^o}, \pi) = \mathcal{M}(\text{\emph{meta-game}}(\boldsymbol{\pi'}))
    \label{equation:delta-uniform-self-play-omega}
\end{equation*}

The functionality of the oracle can be anything that generates a new policy, such as an RL algorithm or evolutionary algorithm amongst other options. Upon completion of the oracle function, a new policy is added to the meta-game. To this extent, the oracle $\mathcal{O}$ and our curator function $G$ are analogous in so far as both functions decide when a policy is introduced in the menagerie. The curator has the advantage of also being able of remove policies from the menagerie.

The extent to which PSRO and our framework overlap is left for future work.

\section{Proposed Incremental Innovations}\label{section:novel-contributions}

% If we keep delta limit uniform, add this back in
In this section we present a novel policy sampling distribution that alleviates on the shortcomings of the $\delta$-Uniform sampling distribution and a novel qualitative metric for the efficiency of the menagerie when it comes to using it as a proxy to the whole policy space. This shows how minimal incremental changes to existing methods, within the context of a general framework, can lead to improvements. 

\begin{comment}
In this section we present a novel qualitative metric that displays how effective a menagerie is at acting as a proxy for the whole policy space. This metric is designed as a tool to observe the interactions between agents in an environment.
\end{comment}

\subsubsection{$\delta$-Limit Uniform policy sampling distribution}
In supervised learning approaches, training datasets are fixed before training commences. This yields a stationary distribution from which training examples are drawn. RL suffers from sequential and correlated data collection during training, rendering a non-stationary distribution over training samples.

We analyze a property of the $\delta$-Uniform SP algorithm. As stated earlier, it aims to generate an agent which can defeat \textit{random} versions of itself. However, this is affected by the sequential data collection curse of RL methods. By sampling uniformly at random from a menagerie, we observe a bias of the policies sampled from $\Omega$ towards earlier policies. Intuitively, earlier policies are sampled more often by virtue of being electable to sampling more times than recently added policies.
\begin{figure}[!b]
\centering
\includegraphics[width=0.35\textwidth, keepaspectratio]{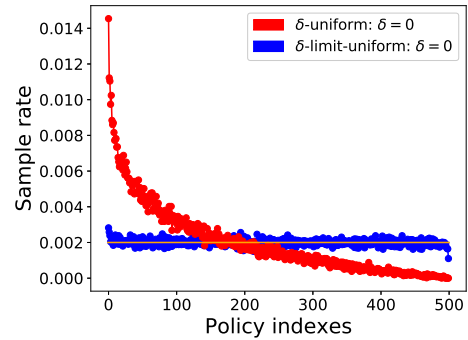}
\caption{Histograms of sample rates for policies inside a menagerie for two sample training runs. The horizontal orange line represents a $Uniform(0, 500)$ distribution.}
\label{fig:delta_vs_limit}
\end{figure}
Computing a policy which generalizes against a broad set of policies is desirable. However, we worry that by sampling earlier policies too often the learning policy will be biased towards interacting with, often random, initial agents. This worry is furthered by empirical evidences stating that, in certain board games, the quality of the fixed policies being used during training is directly proportional to potential quality of the policy being trained~\cite{VanDerRee2013}.

With this in mind, we present a novel policy sampling distribution, named $\delta$-Limit Uniform, that gives increased probability to later policies. An attempt to amend the $\delta$-Uniform bias. Figure~\ref{fig:delta_vs_limit} shows the histograms of the number of samples per policy for both $\delta=0$-Uniform and $\delta=0$-Limit Uniform, clearly showing how the $\delta$-Limit Uniform distribution avoids biasing towards earlier policies.

Let $|\boldsymbol{\pi^o_n}|$ be the size of the menagerie at the beginning of the $n$-th episode. $\pi_e$ is the $e$-th policy to have entered the menagerie (asserting $e \leq n$). The logit probability $\rho_e^n$ and normalized probability $p_e^n$ of sampling $\pi_e$ for the $n$-th SP episode are computed as
\noindent\begin{minipage}{.5\linewidth}
\begin{equation}
    \rho_e^n = \frac{1}{|\boldsymbol{\pi^o_n}|(|\boldsymbol{\pi^o_n}|-e)^2},
    \label{equation:limit-uniform-logit}
\end{equation}
\end{minipage}%
\begin{minipage}{.5\linewidth}
\begin{equation}
    p_e^n = \frac{\rho_e^n}{\sum_{i=0}^{|\boldsymbol{\pi^o_n}|} \rho_i^n}.
    \label{equation:limit-uniform-prob}
\end{equation}
\end{minipage}

\vspace{1em}
\subsubsection{Qualitative Metric for the Menagerie's Efficiency}

A visual metric, aimed at understanding how well a menagerie approximates the entire policy space. Policies can be characterised by the behaviours/state trajectories they produce when acting in the multi-agent environment. Thus, assessing the span of the state trajectories induced by the SP training enables an assessment of the span of the policies living inside the menagerie, which is what we mean by assessing how well a menagerie approximates the whole policy space. This visual display comes from a 2D embedding of the state trajectories experienced by an agent during each training episode. We use t-SNE~\cite{Maaten2008} to project the multi-dimensional, environment specific representation of state trajectories unto a 2D space. Other dimensionality reduction algorithms can be used. We propose two visual cues:
\begin{itemize}
    \item{ 
            \textbf{Density Heightmap}: visualization of the density function yielded by the embedded state trajectories, computed via a kernel density estimation method. Intuitively, it gives insight towards understanding where, inside the embedded state trajectory space, the agent has spent most time on during training. It is valuable providing we can label some subsets of the embedding space with high-level understanding of what is happening throughout the state trajectories.  
    } 
    \item{ 
            \textbf{Time Window-Avegared SP induced trajectories}: visualization of the temporal evolution of the average embedded trajectory/episode for an agent during training. Computed by uniformly dividing the time-sorted embedded trajectories in buckets, with the window-averaged trajectory being the median trajectory, computed in the 2D embedding space, of each bucket. Intuitively, it displays which parts of the embedded trajectory space the agent has traversed throughout training. This cue can be used to visually assess to what extent an agent is prone to re-visit some areas of the trajectory space, which can help identify catastrophic forgetting and cyclic policy evolutions.
    }
            % visualization as a temporal trajectory in the embedding space of the average embedding of SP game trajectories that have been experience by the SP agent in each given time window. Intuitively, it displays which parts of the embedded trajectory space the agent has traversed during training, and how this evolves over time. This cue can be used to visually assess to what extent an agent is prone to re-visit some areas of the trajectory space, which can help identify catastrophic forgetting and cyclic policy evolutions.
\end{itemize}

t-SNE projected representations vary depending on the data used as input. For our purposes it means that if we were to separately embed two sets of different state trajectories, we might not be able to meaningfully compare both separate embeddings. We tackle this problem with two measures: \\
(1) We compute a basis of possible state trajectories using some environment-specific heuristics that enables the basis to span over most of the whole state trajectory space. The number of basis state trajectories computed is of the same order as the number of state trajectories generated during training. 
%To our knowledge, computing a basis of trajectories is an environment specific task.
(2) When comparing two or more sets of state trajectories generated by different algorithms, we compute the embeddings of each algorithm-induced state trajectories all at once via an \textit{aggregated} set of state trajectories. Thus, it allows for meaningful comparisons across state trajectory embeddings from different algorithms.

\section{Experimental Details}\label{section:experimental-details}

\begin{comment}
As it was suggested in~\cite{VanDerRee2013} and~\cite{Firoiu2017}, the benefit of naive SP was dependant on which algorithm was used to train the learning policy. We aim to shed new observations on this matter, and to extend these experiments to other SP schemes. We aim to evaluate the effect of different SP training schemes on various algorithms. as a measure of how policies trained using these algorithms perform against a set of fixed opponents. Note that these fixed opponents are used only during testing and not during training, as otherwise we would be breaking Assumption 1 from Section~\ref{subsection:self-play-assumptions}. 
\end{comment}

% In this section, we aim to quantitatively evaluate the effect of different SP training schemes when using PPO~\cite{schulman2017proximal}, a popular on-policy RL algorithm.

\subsection{Experiment description}

We now present the environments, evaluation metrics, RL algorithms and SP schemes used in our experiment.

\subsubsection{Environment}

Repeated imperfect Recall Rock Paper Scissors (RirRPS) as introduced in~\cite{Hernandez2019}. An extended form, imperfect imformation, two-player, zero-sum, simultaneous version of Rock Paper Scisors. The agent which obtains the highest cumulative reward by the end of the last repetition is considered the winner. Ties are broken uniformly at random. We choose a \textit{repeated} game and not a single round because repeatability introduces explotability, which increases with the number of repetitions. In our experiments we use 10 repetitions, with a recall of the last 3 joint actions.

% \paragraph{Kuhn poker~\cite{kuhn1950simplified}} 
% 
% An extremely simplified form of poker; A zero-sum, two-player, imperfect-information, sequential game. In Kuhn poker, the deck includes only three playing cards, for example a King, Queen, and Jack. One card is dealt to each player, which may place bets similarly to a standard poker. If both players bet or both players pass, the player with the higher card wins, otherwise, the betting player wins.

RirRPS is a highly (but not fully) cyclic game. In fully cyclic games, improving an agent's performance against another agent is always counterbalanced by a decrease in performance against other possible agents, implying that invididual agent improvement is inconsequential~\cite{Balduzzi2019}.

%In contrast, Kuhn's poker is a highly transitive game, meaning that there are certain strategies which are objectively worse than others. We want to explore how different SP schemes react to both cyclic and transitive environments.

\subsubsection{Algorithmic choices}

\begin{table}[h]
    \centering
    \caption{PPO hyperparameters used for both experiments.}
    \begin{tabular}{  l  c  c }
        \toprule
        Hyperparameter & Qualitative study & Quantitative study\\
        \midrule
        Horizon (T) & $2048$ & 128 \\
        Adam stepsize & $3\times10^{-4}$ & $10^{-5}$ \\
        Num. epochs & $10$ & 10 \\
        Minibatch size & $64$ & 16 \\
        Discount ($\gamma$) & $0.99$ & $0.99$ \\
        GAE parameter ($\lambda$) & $0.95$ & $0.95$ \\
        Entropy coeff. & $0.01$ & $0.01$ \\
        Clipping parameter ($\epsilon$) & $0.2$ & $0.2$ \\
        \bottomrule
    \end{tabular}
    \label{table:PPOHyper}
\end{table}

For our qualitative studies we used Proximal Policy Optimization~\cite{schulman2017proximal} where the underlying policy is represented by either by a feedforward neural network (MLP-PPO) or a recurrent architecture (RNN-PPO). Four our quantitative studies we only use MLP-PPO.

% The underlying RL algorithm used to update agent's policies is the Proximal Policy Optimization (PPO) algorithm~\cite{schulman2017proximal}. Hyperparameters are presented in table~\ref{table:PPOHyper}.

\subsubsection{Self-Play choices}

We train a PPO agent on a SP extended MARL loop as shown in Algorithm~\ref{algorithm:posg-self-play-rl-loop}:
\begin{itemize}
    \item Naive SP
    \item $\delta$-Uniform and $\delta$-Limit Uniform, where the value of $\delta$ is specified each time.
    \item PSRO($\mathcal{M}=$ maxent-Nash, $\mathcal{O}=$ Best Response). Such oracle is governed by two hyperparameters, which play a role in determining whether the training agent has converged to a best response: (1) The winrate $w \in [0,1]$ at which it is considered that the current agent has conveged and (2) the number of episodes $n_{matches}$ that will be used to compute the aforementioned winrate. We used $w=72\%$, $n_{matches}=50$.
\end{itemize}

For all SP training schemes, the initial menagerie contains a copy of the initial policy, with randomly initialized weights.

\subsubsection{Evaluation metrics}

\paragraph{\textbf{Winrate matrices}}

SP algorithms train / modify a policy $\pi$ overtime. We can consider an SP scheme $sp$ as a generative process, which we can query at any time $t$ to obtain the latest version of $\pi$ being trained by $sp$, $\pi_t \sim sp$. This is analogous to creating checkpoints in training at which to freeze a copy of the policy $\pi$ being trained. We only freeze $\pi_t$ and not the menagerie $\bm{\pi^o_t}$. Thus, we can generate a population which represents the evolution of the policy training under an SP algorithm overtime, $\bm{\pi_{sp}} = [\pi_{t_0}, \pi_{t_1}, \dots]$. By examining the evaluation matrix generated from this population $\bm{W_{\pi_{sp}}}$ we can quantitatively examine if different SP algorithms suffer from catastrophic forgetting or cyclic policy evolutions.

% We use quantitative meassures of performance based on evaluation matrices~\cite{Balduzzi2019} favoured in recent research~\cite{vinyals2019grandmaster}~\cite{liu2019emergent}. These meassures are based on computing empirical winrate matrices as described in Section~\ref{section:preliminary_notation}. The population used 
% 
% \begin{enumerate}
%     \item Sample a sequence of $N$ policies from SP training scheme. $\{\pi\}_{1 \ldots N} \sim sp$
%     \item Compute empirical winrate matrix $W \in \mathbb{R}^{N \times N}$ from the sequence of sampled policies. The entry $w_{ij}$ for $i,j \in \{1 \ldots N\}$ represents the winrate of policy $\pi_i$ against $\pi_j$ for a given number of episodes. Both $\pi_i , \pi_j \in \{\pi\}_{1\ldots N}$
%     \item Compute all submatrices of $W$, $W_{sub} = \{ W_{1 \ldots i \times 1 \ldots i}: i \in \{2 \ldots N\}\}$. Note that $W$ is contained in its set of submatrices $W \in W_{sub}$.
%     \item Compute Nash Averaging  (Balduzzi, 2019) for all matrices in $W_{sub}$. Let the nash averaging for matrix $W_i \in W_{sub}$ be denoted by $\text{\textbf{n}}_{W_{i}}$.
% \end{enumerate}

\paragraph{\textbf{Evolution of relative population performance}}

As introduced in Section~\ref{section:preliminary_notation}, we shall use the relative population performance as a direct meassure of the relative quality between the populations spawned by two different SP algorithms. We are interested in how this relative performance evolves overtime. Below we describe the algorithm to obtain such evolution: Given a set of SP training algorithms $SP$:
\begin{enumerate}
    \item For each $sp \in SP$ sample a population $\bm{\pi_{sp}}$ of $n$ agents.
    \item For each population pair $(\bm{\pi_{sp_1}},\bm{\pi_{sp_2}}), \; sp_1, sp_2 \in SP$, compute an evaluation matrix $\bm{A_{\pi_{sp_1}, \pi_{sp_2}}}$ between both populations.
    \item Compute $\bm{A_{sub}} = \{ \bm{A_{1 \ldots i \times 1 \ldots i}}: i \in \{n\}\}$, which represents all submatrices of $A_{\bm{\pi_{sp_1}}, \bm{\pi_{sp_2}}}$.
    \item Compute the evolution of relative population performance associated with each submatrix $\bm{A_i} \in \bm{A_{sub}}$, $\bm{v_{sp_1, sp_2}} = [v_{\bm{A_i}}] \in \mathbb{R}^n$.
\end{enumerate}

Evaluation matrices are expensive to compute: $O(n^2)$ where $n$ is the population size. There is current research on reducing the computational load of generating evaluation matrices~\cite{rowland2019multiagent}. The procedure outlined above uses a single evaluation matrix to compute the relative population performances for all submatrices, meaning that we can recycle the empirical winrate matrix used to generate the evaluation matrices. Throughout this paper, to compute the winrate for an entry $w_{i,j}$ in an empirical winrate matrix $\bm{W}$ we use 30 simulations.

\section{Results}\label{section:results}

\subsection{Qualitative analysis}

\begin{figure*}[!t]
    \centering
    \includegraphics[width=.8\textwidth]{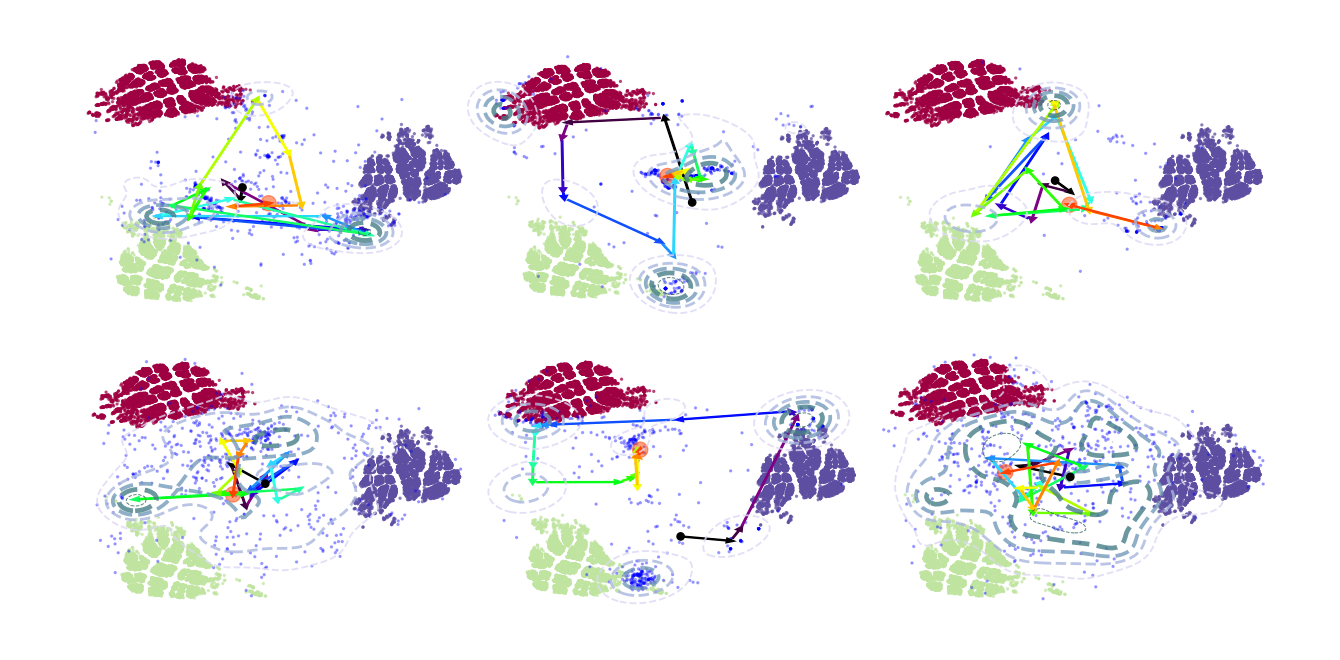}
    \caption{Density Heightmap and Time Window-averaged SP-induced of episode trajectories in the computed 2D t-SNE state trajectory embedding space. \textbf{Top-Left:} Naive SP with MLP-PPO. \textbf{Bottom-Left:} Naive SP with RNN-PPO. \textbf{Top-Centre:} $\delta=0$-Uniform SP with MLP-PPO. \textbf{Bottom-Centre} $\delta=0$-Uniform SP with RNN-PPO. \textbf{Top-Right:} $\delta=0$-Limit Uniform SP with MLP-PPO.  \textbf{Bottom-Right:} $\delta=0$-Limit Uniform SP with RNN-PPO. RirRPS environment, $1e4$ SP training episodes. Green, purple, and red-colored clusters are embeddings of state trajectories resulting of pitting, respectively, RockAgent, PaperAgent, and ScissorsAgent against a RandomAgent. The scattered blue dots represents the individual projection of each one of the $10e4$ trajectories. Their density heightmaps are represented through dashed contours. The time-sorted training trajectories experienced by the SP agents were divided into $20$ time-windows, and a centroid (median trajectory) was computed for each. Consecutive centroids have been linked by arrows, creating the Time Window-averaged SP-induced episode trajectories. Starting at the black dot, their progression is highlighted via the rainbow colour transitions.}
    \label{fig:tsne}
\end{figure*}

Figure~\ref{fig:tsne} shows the $2$D t-SNE state trajectory embeddings for all combinations of SP algorihtm \& RL algorithm introduced in the previous section. Each training session lasted for a $1e4$ episodes on the RirRPS environment. 

Each SP agent using naive SP and $\delta=0$-Limit Uniform exhibits cyclic catastrophic forgetting as their time window-averaged trajectories in the embedded space display cyclic movement, whereas $\delta=0$-Uniform's time window-averaged trajectories seem less affected.

Especially in the cases of the $\delta=0$-Limit Uniform and Naive SPs, the Density Heightmaps of RNN-PPO seem to be made of plateaus whereas the ones of MLP-PPO  are made of picks, indicating that recurrent policies seem to further spread the menagerie over the whole policy space to some greater extent compared to feedforward policies.

Comparing $\delta=0$ Uniform and $\delta=0$-Limit Uniform SPs, we can observe a progressive and somewhat ordered exploration of the policy space by the former. $\delta=0$-Uniform's Time Window-averaged SP episode trajectories visit each fixed agent clusters one by one. Since the former biases towards earlier policies that have entered the menagerie when sampling opponent, we hypothesize that this time-related bias is entering in synergy with the learning rate of the trained policy. Indeed, after behaving like a Rock Agent (green cluster), the trained policy starts to behave like a Paper Agent (purple cluster) as the Rock Agent-behaving policies that have entered in the menagerie progressively starts to be sampled as opponent. Both the MLP-PPO- and RNN-PPO-equipped agents exhibit that cyclic and ordered exploration of the embedding space.

\subsection{Quantitative analysis}

\begin{figure*}[h]
    \centering
    \includegraphics[width=\textwidth, keepaspectratio]{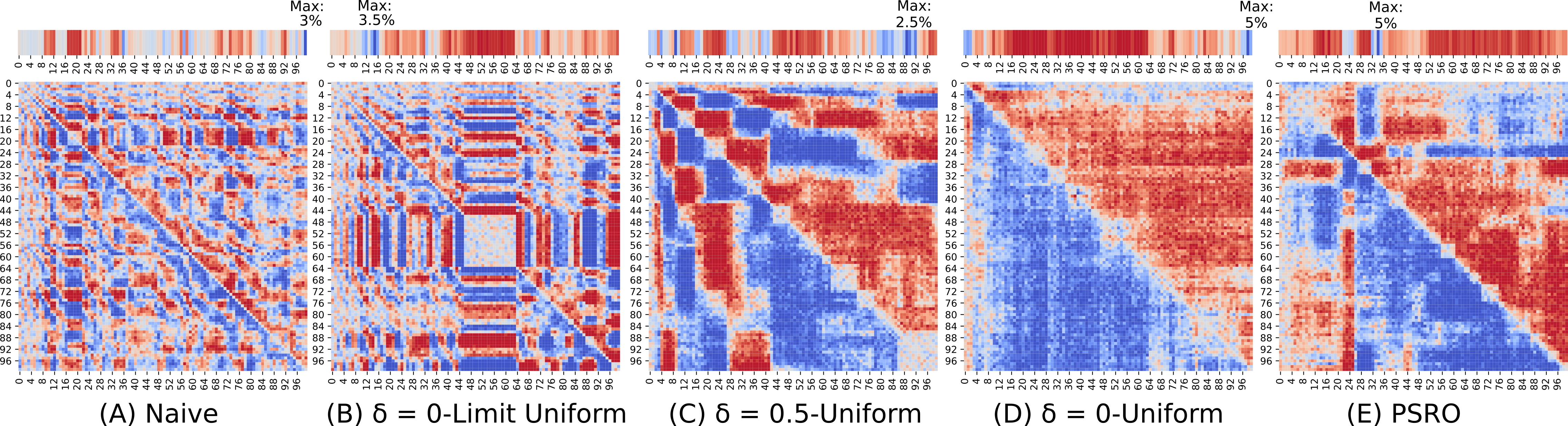}
    \caption{Empirical winrate matrices showing the evolution of 6 policies where each one is being trained via a different SP algorithm in RirRPS. For every SP training process, we sample a policy after every policy update for a total of 100 policy checkpoints and 12800 training episodes. Treating each matrix as the payoff matrix for a symmetrical 2-player zero-sum game, we present on top of of each matrix the support received by each policy on the Nash equilibrium of such game. This support gives a measure of quality of each individual policy with respect to the other policies in the population. \textbf{Blue} / \textbf{red} indicates \textbf{positive} / \textbf{negative} winrates for column player.}
    \label{figure:winrate_matrices_all_sp}
\end{figure*}

The results from Figure~\ref{figure:winrate_matrices_all_sp} are metrics gathered on a single training run due to the computational requirements of averaging results over many runs. However, the behaviour captured is a representative sample of many training runs.

% Reinforcement learning notoriously suffers from variance~\cite{Lanctot2017}\cite{Henderson2017}.

Each row $i$ of winrate matrix $\bm{W}$ represents the winrates of policy at checkpoint $i$ against all other policies checkpointed during training. Thus, for any given row $i$, the entries left of the diagonal ($w_{i,j}, \forall j < i$) indicate winrates against policies from earlier checkpoints in training, or older policies. Conversely, entries right of the diagonal ($w_{i,j}, \forall j > i$) denote winrates of policy $i$ against later checkpoints, or newer policies. Diagonal entries represent the winrate of a policy against itself, which is always 50\%. An ideal training scheme which would always compute monotonically better policies as training progressed would yield a winrate matrix where the lower triangular indices would show positive winrates (higher than 50\%) and the upper triangular would show negative winrates (lower than 50\%). In other words, a policy would always win against previous versions of itself, and lose against newer ones.

We turn our focus to the winrate matrices from Figure~\ref{figure:winrate_matrices_all_sp}. As discussed, Naive SP uses as opponent an identical version of the policy being trained, and thus the underlying RL algorithm tries to compute a best response against itself. This is clearly manifested in the winrate matrix in Figure~\ref{figure:winrate_matrices_all_sp}.A. The entries just left of the diagonal show positive winrates, and those just right of the diagonal show negative winrates. This means that the training policy learns how to beat the last version of itself.

Figure~\ref{figure:winrate_matrices_all_sp}.C shows the evolution of the policy training under $\delta = 0.5$-Uniform (Half history). This policy attempts to compute a best response against the later half of its history. We see that on average, for a given row $i$, the corresponding policy tends to win against policies $j \in [\frac{i}{2}, i-1]$. Note that policies immediately outside the moving window determined by the choice of $\delta = 0.5$ feature a negative winrate, suggesting the training policy does not generalize to policies outside of the menagerie in RirRPS.

$\delta = 0$-Uniform (Figure~\ref{figure:winrate_matrices_all_sp}.D), whose underlying policy attempts a best response against its entire history, shows a close-to-ideal empirical winrate matrix insofar as any given policy $i$ beats most previous versions of itself and loses against later ones. Also, for any subgame of Figure~\ref{figure:winrate_matrices_all_sp}.D the largest concentration of support under Nash consistently lays on the latest policies.

PSRO's winrate matrix, depicted in Figure~\ref{figure:winrate_matrices_all_sp}.E, does follow a positive trend, although less so than $\delta=0$-Uniform, as checkpoints beyond the 37th lose against policies 20 to 26, which worsens as later policies are introduced. Interestingly, the policy featuring the largest support under Nash is the 34th checkpoint. This does \textit{not} necessarily mean that all 66 checkpoints that came after it were weaker in comparison. The quality of a policy (in terms of support under Nash) can vary greatly when policies are added or dropped from the population. For instance, if we consider a subgame of Figure~\ref{figure:winrate_matrices_all_sp}.E taking only the first 92 checkpoints, we would find that the 92nd policy features the largest support under Nash around 3\%, yet it falls around 1\% on the game from Figure~\ref{figure:winrate_matrices_all_sp}.E.

\section{Discussion}

\paragraph{\textbf{Cyclic policy evolutions}}

As expected, naive SP clearly features a cyclic policy evolution. As previously stated, an ideal SP would yield policies that always beat previous ones. In contrast, almost all checkpoints obtained during naive SP training cycle between losing and winning against previous and future checkpoints. This is further evidenced by the support under Nash from Figure~\ref{figure:winrate_matrices_all_sp}.A, where under Nash equilibrium many policies share the highest amount support (around 3\%). We observed similar cyclic behaviour in $\delta=0$-Limit Uniform in Figure~\ref{figure:winrate_matrices_all_sp}.B and in $\delta=0.5$-Limit Uniform (not shown). Which may entail that $\delta$-Limit Uniform SPs over-correct the bias towards earlier policies, matching our observations on the previous qualitative analysis.

$\delta = 0$-Uniform (Full history) does not exhibit a cyclic policy evolution. Full history tends towards generating monotonically better policies. That is, on average any given checkpoint is able to obtain positive winrates when matched against previous checkpoints. Hence, we claim that full history SP is an apt SP scheme for cyclic environments, given enough computational time. On the other hand, policies trained via $\delta=0.5$-Uniform SP seem to struggle to reliably defeat all previous versions of themselves. Hence in RirRPS we see that by excluding the earlier half of the history, the latest policy becomes exploitable by earlier policies. Note that half history's winrate matrix features more pronounced winrates (entries are either closer to 100\% or 0\% winrate) than the equivalent entries in full history's winrate matrix. This is a result of $\delta = 0.5$-Uniform's menagerie being smaller than $\delta = 0$-Uniform's counterpart, which leads the training policy to overfit against the policies in the menagerie, which in turn allow earlier policies to exploit it.

PSRO (Figure~\ref{figure:winrate_matrices_all_sp}.E) does show signs of forgetting, as stated in the previous section. As training progresses, later policies begin to lose against previous ones, with this effect growing larger overtime.

\paragraph{\textbf{Relative population performances}}

\begin{figure}[h]
    \centering
    \includegraphics[width=0.5\textwidth, keepaspectratio]{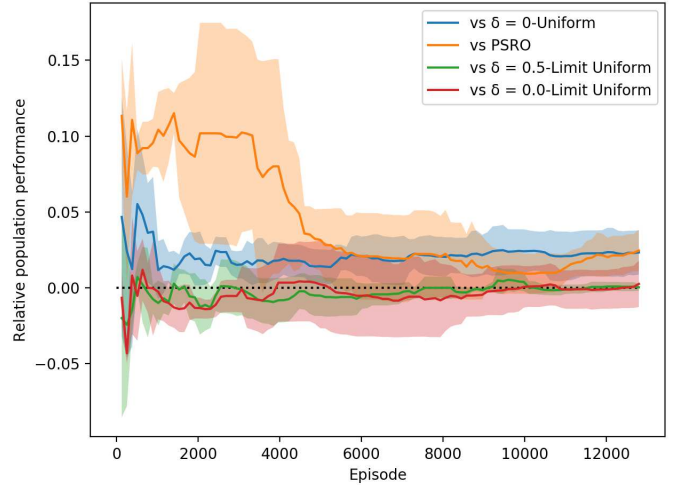}
    \caption{Evolution of relative population performance of naive SP against 4 other SP schemes. The cyclic behaviour of naive SP quickly discovers how to play Rock, Paper and Scissors, which are enough to generate a Nash equilibrium, which explains the initial positive relative performance.}
    \label{figure:relative-population-performance}
\end{figure}

Figure~\ref{figure:relative-population-performance} shows the evolution of population performances comparing the performance of naive SP against the SP algorithms from Figure~\ref{figure:winrate_matrices_all_sp}. Interestingly, when we look at the evolution of relative population performance we notice that it converges near zero for all SP algorithms. For RirRPS, this implies that the populations generated by all SP as training progresses are of similar quality, furthering the idea that in highly cyclic games individual policy improvement is not meaningful, even when there is potential to exploitation due to repetitions in RirRPS. However, we are surprised to find naive SP performing better than $\delta=0$-Uniform and PSRO, which is not obvious by just looking at the winrate matrices from Figure~\ref{figure:winrate_matrices_all_sp}.

A possible reason why naive SP performs evenly or positively against all other SP algorithms is that early on in training it quickly cycles through rock / paper / scissors policies, and from those three policies it is possible to compose almost any policy in RirRPS.

\paragraph{\textbf{Fragility of PSRO's oracle hyperparameters}}

Small changes in the oracle's hyperparameters (winrate threshold $w$, window size of match outcomes $n_{matches}$) can quickly lead to unfeasibly long training times (too many policies added to the menagerie) or arguably degenerate behaviour by the SP algorithm (the curator never introduces new policies into the menagerie). In the worst case scenario, the training policy will never convege towards a best response against the initial (randomly initialized) policy in the menagerie. We show in Table~\ref{table:PSRO-sweep} a sweep over both hyperparameters in RirRPS. Most of the training time is spent inside of the meta-game solver $\mathcal{M}$, which leads us to believe that a less computationally intensive meta-game solver should be used.

A Nash Equilibrium in RirRPS is to act randomly, which we argue is likely the behaviour of the training policy at the beginning of training. Hence, it is highly unlikely that a policy will obtain a high enough winrate against this random policy to be added to the menagerie, making it difficult for the learning policy to discover policies which differ from random play. This means that the policy will not discover how to exploit policies beyond random play.

\begin{table}[h]
    \centering
    \caption{Hyperparameter sweep time profiling for 12k episodes in RirRPS. Columns $\mathcal{M}$ and $\bm{W_{\pi^o}}$ represent the percentage of training time spent on computing a meta-game solution and updating the meta-game respectively. }
    \begin{tabular}{c c c c c}
        \toprule
        Hyperparameter values & $\mathcal{M}$ & $\bm{W_{\pi^o}}$ & Total training time & $|\bm{\pi^o}|$\\
        \midrule
        (60\%, 30) & 95\% & 2\% & $>$2d & 332 \\
        (70\%, 30) & 95\% & 5\% & 1d 12h 44m & 294 \\
        (75\%, 30) & 70\% & 25\% & 1h 38m & 139 \\
        (80\%, 30) & 59\% & 32\% & 55m & 118 \\
        (85\%, 30) & 0\% & 0\% & 4m & 1 \\
        (70\%, 45) & 69\% & 24\% & 58m & 112 \\
        (75\%, 45) & 2\% & 9\% & 5m & 19 \\
        (80\%, 45) & 0\% & 0\% & 4m & 1 \\
        (70\%, 50) & 67\% & 26\% & 1h 7m & 123 \\
        \bottomrule
    \end{tabular}
    \label{table:PSRO-sweep}
\end{table}

\paragraph{\textbf{Extrapolation}}
RirRPS is a \textit{2-player}, \textit{zero-sum} and \textit{simultaneous} game. Our experimental results may not extend to \textit{n-players} or \textit{general-sum} games. Moreover, following the dimensionality-based definition of the complexity of a game \cite{Balduzzi2019}, the lower-bound on the complexity of RPS and, incidentally, RirRPS are rather low. Therefore, it would be interesting to compare current results with games of verifiably greater lower-bound on their complexity, such as RoboSumo \cite{alshedivat2017continuous}.

\section{Conclusions \& Future Work}\label{section:conclusion}

Building on our original work~\cite{Hernandez2019}, this paper presents a general framework in which to define SP training schemes. This is done by formalizing the notion of a menagerie, a policy sampling distribution and a curator (gating) function. This framework is framed as theoretical approximation to a solution concept in MARL\@, under stated assumptions. The framework's generalizing capabilities have been showcased by capturing existing SP algorithms within it. We have also identified shortcomings of some of the captured methods, and have proposed methods which could potentially overcome said issues. Through a qualitative study we have showcased that, on a simple environment, different SP algorithms differ in how the joint policy space is explored. We have also carried out a quantitative analysis on (1) the evolution of policies being trained under different SP algorithms to discover cyclic policy evolutions and (2) the relative performance between various SP algorithms.

Future work will study other possibilities presented within the expressive capabilities of our SP framework. For instance, there is no research exploring which policy sampling distribution works best for different types of environments. Furthermore, it may even be possible to \textit{learn} a policy sampling distribution or curator during training using meta RL\@.

\section*{Acknowledgements}

We deeply appreciate Jayesh K. Gupta for his insightful conversations and work on Nash averaging.

% This work was funded by the EPSRC Centre for Doctoral Training in Intelligent Games and Game Intelligence (IGGI) EP/L015846/1.
\vspace{-0.5em}

\bibliographystyle{IEEEtran}
\bibliography{IEEEabrv,main}

\end{document}